\def\year{2022}\relax
\documentclass[letterpaper]{article} 
\usepackage{aaai22}  
\usepackage{times}  
\usepackage{helvet}  
\usepackage{courier}  
\usepackage[hyphens]{url}  
\usepackage{graphicx} 
\usepackage{natbib}  
\usepackage{caption} 
\DeclareCaptionStyle{ruled}{labelfont=normalfont,labelsep=colon,strut=off} 
\usepackage{algorithm}
\usepackage{algorithmic}
\usepackage{amsmath}
\usepackage{newfloat}
\usepackage{listings}
\usepackage{amssymb}
\usepackage{float}
\graphicspath{{pic/}}
\floatstyle{ruled}
\newfloat{listing}{tb}{lst}{}
\floatname{listing}{Listing}
\title{MAGIC: Multimodal relAtional Graph adversarIal inferenCe \\for Diverse and Unpaired Text-based Image Captioning}

\author {
Wenqiao  Zhang $\textsuperscript{\rm 1}$$\footnote{These authors contributed equally to this work.}$,
        Haochen Shi $\textsuperscript{\rm 2}$$\footnotemark[\value{footnote}]$,
        Jiannan Guo $\textsuperscript{\rm 1}$$\footnotemark[\value{footnote}]$,
        Shengyu Zhang $\textsuperscript{\rm 1}$
        Qingpeng Cai $\textsuperscript{\rm 3}$, \\
        Juncheng Li $\textsuperscript{\rm 1}$$ \footnote{Corresponding author.}$,
        Sihui Luo $\textsuperscript{\rm 4}$,
       Yueting Zhuang $\textsuperscript{\rm 1}$$\footnotemark[\value{footnote}]$\\
}

\affiliations {
    $\textsuperscript{\rm 1}$ Zhejiang University,
    $\textsuperscript{\rm 2}$ Universit\'{e} de Montr\'{e}al,
    $\textsuperscript{\rm 3}$ National University of Singapore,
    $\textsuperscript{\rm 4}$ Ningbo University\\
    \tt\small {$\{$wenqiaozhang, jiannan, sy$\_$zhang junchengli, yzhuang$\}$@zju.edu.cn},
    \tt\small {haochen.shi@umontreal.ca, qingpeng@comp.nus.edu.sg, luosihui@nbu.edu.cn}
}

\usepackage{bibentry}

\begin{document}
\maketitle

\begin{abstract}
Text-based image captioning (TextCap) requires simultaneous comprehension of visual content and reading the text of images to generate a natural language description. Although a task can teach machines to understand the complex human environment further given that text is omnipresent in our daily surroundings, it poses additional challenges in normal captioning. 
 A text-based image intuitively contains abundant and complex multimodal relational content, that is, image details can be described diversely from multiview rather than a single caption.  Certainly, we can introduce additional paired training data to show the diversity of images' descriptions, this process is labor-intensive and time-consuming for TextCap pair annotations with extra texts. Based on the insight mentioned above, we investigate how to generate diverse captions that focus on different image parts using an unpaired training paradigm. We propose the Multimodal relAtional Graph adversarIal inferenCe (MAGIC) framework for diverse and unpaired TextCap. This framework can adaptively construct multiple multimodal relational graphs of images and model complex relationships among graphs to represent descriptive diversity. Moreover,  a cascaded generative adversarial network is developed from modeled graphs to infer the unpaired caption generation in image–sentence feature alignment and linguistic coherence levels. We validate the effectiveness of MAGIC in generating diverse captions from different relational information items of an image. Experimental results show that MAGIC can generate very promising outcomes without using any image–caption training pairs.

\end{abstract}

\begin{figure}[t]
\includegraphics[width=0.5\textwidth]{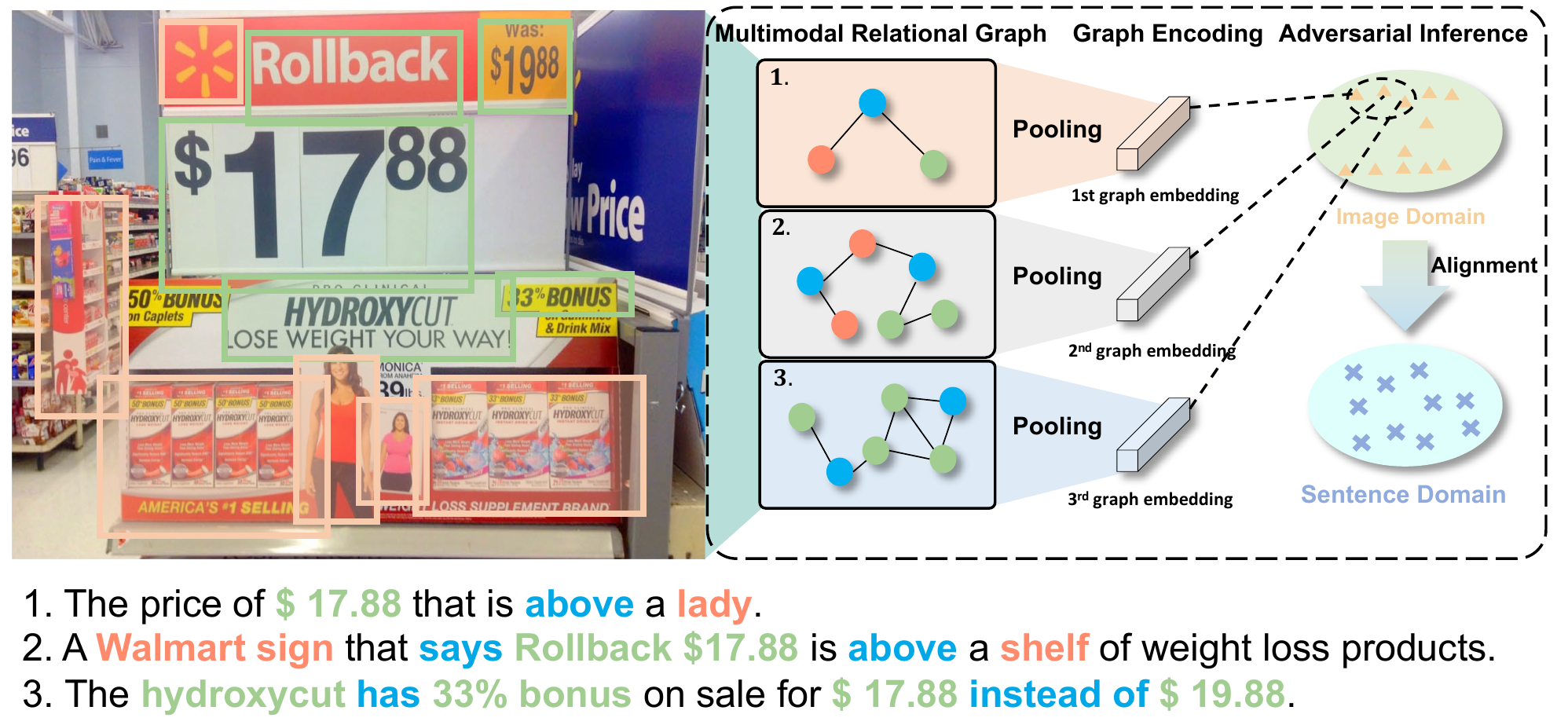}
\centering\caption{  An  example  of  how  the diverse and unpaired captions generated. Orange, green and blue represent the objects, texts and multimodal relationships, respectively.}
\vspace{-0.2cm}
\label{fig1}
\end{figure}

\section{Introduction}
Visual captioning aims to help visually impaired people to
quickly understand the image content. Existing methods focus on describing visual objects that have achieved impressive or even super-human performance~\cite{yu2019multimodal,cornia2020meshed,zhang2021consensus,ding2020stimulus,zhang2021consensus,zhang2021tell}. However,  most prominent captioning methods fail to ``read'' the associated texts in the visual scene.  Therefore, text-based image captioning (TextCap)~\cite{sidorov2020textcaps} is proposed in this study to caption models, recognize visual content, read text, understand both modalities jointly, and write a succinct caption. Such task is very practical that encourage machine to generate fine-grained captions (e.g., read the price of commodities in Fig.\ref{fig1}), which is helpful to visually disabled people to understand surroundings more accurately, but put forward new difficulties to existing captioning models meanwhile.


Intuitively, human-made images with text contain more semantic entities (e.g., objects, attributes, and texts) and complex relations (e.g., correlation and constraint) than natural images~\cite{hou2019relational,zhang2020relational,zhang2020photo}. Common single captioning models experience difficulty in representing a holistic image understanding.  These models also hinder the diversity of image descriptions, which is crucial for blind people concerning different aspects of an image (e.g., price comparison, discount, brand and position of commodities in Fig.1). In fact, we humans tend to focus on different aspects of image content by our preferences, personality and sentiment, this applies to the annotators with no exception~\cite{yang2020deconfounded}. Definitely, we could a
sk annotators for more paired data (e.g., \emph{multiview of an image} - \emph{diverse descriptions}), thereby force captioning model to generate varied captions from multiple views of an image. However, in the Textcap task as well as many vision-language applications, such large-scale annotations are not readily available, and are both time-consuming and labor-intensive to acquire. In these scenarios, unsupervised/unpaired methods~\cite{lample2017unsupervised,gu2019unpaired,caron2020unsupervised} that can learn vision to language or conversely from unpaired training data are highly desirable. Based on this insight, our goal is to break away from single and paired captioning conventions and conduct the unpaired captioning paradigm, which encourages TextCap models to generate diverse textual descriptions to comprehend images better.




However, the diverse and unpaired TextCap is challenging and can be summarized as three aspects:  1)~\textbf{Descriptive Diversity}, as shown in Fig.1, three sentences focus on different and interrelated multimodal information. It is hard to determine which fine-grained multimodal details are semantically related and ignore the unconcerned content to mine the diversity of image content from multiple views; 2)~\textbf{Relational Modeling}, how to model complex intra- and cross-relationships among the above multimodal contents, some of which even seem irrelevant intuitively. For instance there are various relational semantics in Fig.\ref{fig1}: (a) \emph{intra-relationship},
a price of ``17.8'' denotes the ``rollback'' while ``19.88'' indicates the original price, a lady is located next to the commodity;
 (b) \emph{cross-relationship}, the word “Hydroxycut” is
  the brand of a commodity and its price of “17.88” is located above a lady. Thus, appropriately model the relational semantic meanings is the premise to generate a plausible caption; 3)~\textbf{Unpaired Learning Paradigm}, there have been few attempts for unpaired captioning~\cite{gu2018unpaired,gu2019unpaired}, mainly relies on aligning/sharing latent semantic space across the image-sentence domain. Although promising,  these methods may not mature enough to apply to unpaired Textcap. The multimodal relationships among images are complex and abundant. The characteristics distribution among images and descriptions are apparently larger than general captioning.

 
 

Overall, we propose a novel method for diverse and unpaired TextCap called Multimodal relAtional Graph adversarIal inferenCe (MAGIC)  to address these challenges. As shown in Fig.1, our framework comprises the following parts: 1) \emph{Multimodal Relational Graph Encoder}, this module aims to encode an image to multiple semantic embeddings based on content diversity adaptively.  The intuition behind this is: when human beings composing different image descriptions, they prefer to focus on a few salient regions,  then query the contextually relevant content part and clarify their relationships to describe. To mimic this process, we first introduce a central object-aware pool that adaptively identifies the key objects which are supposed to reflect humans' intentions. Then construct their multimodal relational graphs via query relation-aware multimodal semantic meaning. The relational contexts among graphs are finally modeled by GCN~\cite{berg2017graph} into embeddings to represent varying levels of details of an image;  2) \emph{Sentence Auto-Encoder},  this module aims to reconstruct the sentences by encoding the pre-extracted scene graph from the sentence corpus. Encoded embedding sets can learn the language inductive bias~\cite{yang2020auto} from the sentence corpus to provide explicit signals for subsequent unpaired mapping and decode lexical words appropriately; 3) \emph{Unpaired Adversarial Caption Inference}, last but not least, semantic consistency throughout image and sentence domain is crucial to achieving unpaired learning. For this purpose, we develop a cascaded generative adversarial network(CGAN). CGAN first performs feature alignment from image to sentence to capture the cross-modal semantic consistency, then learns linguistic coherence by a language discriminator to generate more accurate and fluent captions.   To summarize, the major contributions of our paper are as follows:  

\begin{itemize}
\item  A novel perspective is presented for the TextCap task to simultaneously liberate from the single and paired captioning training mode.

\item The MAGIC framework can appropriately construct and encode diverse multimodal relational graphs to observe an image from multiview and learn to generate unpaired captions via cascaded adversarial inference.

\item The superiority of MAGIC is demonstrated in terms of both accuracy and diversity using automatic metrics and human judgments.
 \end{itemize}
\vspace{-0.2cm}

\section{Related Work}

\noindent\textbf{Diverse Captioning.} The ability of understanding and reasoning different modalities  is
a longstanding and challenging goal of artificial intelligence~\cite{rennie2017self,zhang2020photo,zhang2019frame,li2021adaptive,li2020unsupervised,li2019walking,chen2021cross}. Image captioning is one of the most important task and
recent works are focus on generating multiple captions to describe an image.  \cite{johnson2016densecap} is the first to propose the dense captioning task, which detects and describes diverse regions in the image.   ~\cite{guo2019mscap} design a multi-style image captioning model that learns to map images into attractive captions of multiple styles. ~\cite{xu2021towards} develops anchor-centered graphs that select different textual contents to generate that multiple captions. However, the works mentioned above are more relevant to holistic visual understanding by paying more attention to unimodal information (e.g., objects, regions or texts). They require much paired data to learn the captioning diversity and may hardly model an image's complex multimodal relational details. 

\begin{figure*}[t]
\includegraphics[width=0.9\textwidth]{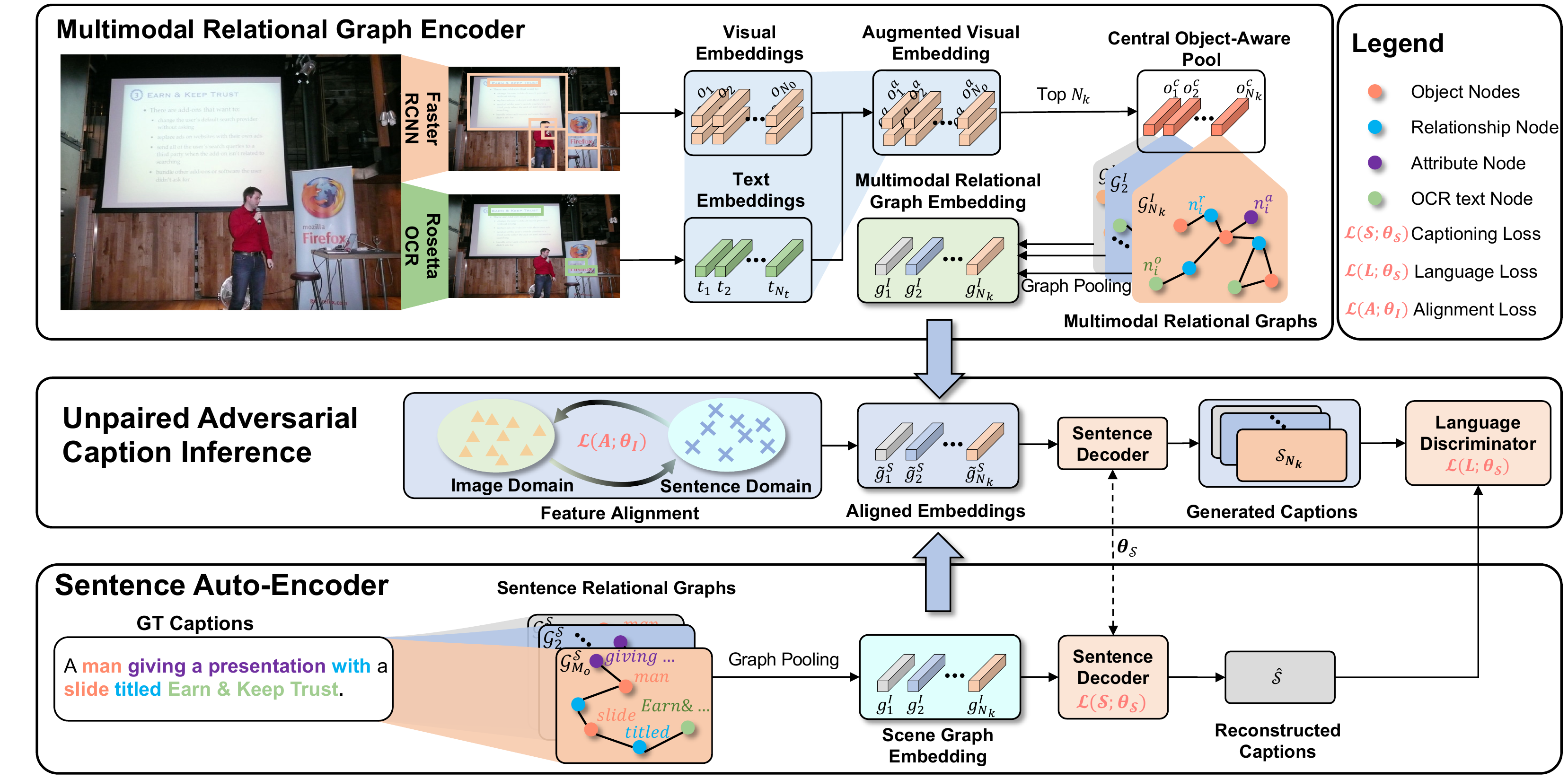}
\centering\caption{Overview of our framework.  (a) \emph{Multimodal Relational Graph Encoder} encodes multiple multimodal relational graphs based on central object-aware pool adaptively to represent the diversity of image content;
 (b) \emph{Sentence Auto-Encoder} pre-extract the scene graph from the sentence corpus,  then reconstructing the sentences from the latent vectors that are encoded from the scene graph; (c) \emph{Unpaired Adversarial Caption Inference}, a cascaded generative adversarial network developed to reason the accurate and fluent captions with unpaired learning by aligning cross-domain features and distinguishing linguistic structure.}
\end{figure*}
\noindent\textbf{Unpaired Captioning.} 
First attempt~\cite{gu2018unpaired} of unpaired image captioning   utilize pivot language corpus and align it to the target language,  connect the pivot language sentences in different domains by shared word embeddings. ~\cite{gu2019unpaired} uses scene graph as representations of the image and the sentence, and maps the scene graphs in their feature space through cycle-consistent adversarial training. The most recent work\cite{liu2021exploring} employs the semantic relationships to bridge the gap between vision and language domains for unpaired caption generation. However, due to the complex multimodal relationships in TextCap task, the discrepancy of feature distribution and semantic gap across image-sentence domain is larger than image captioning. The above methods may fail to generate accurate captions appropriately in an unpaired setting. 

\section{Method}
This section describes our diverse and unpaired TextCap framework, we will describe each module of it and introduce the objectives and strategy for training.
\subsection{Problem formulation}
Before presenting our method, we first introduce some basic notions and terminologies. Given a text-based image $\mathcal{I}$,  the goal is to generate diverse natural language sentences $\mathcal{S}=\{\mathcal{S}_i|_{i=1}^{N_k}\}$ with unpaired learning, where $N_k$ is the pre-defined number of generated sentences to satisfy the user’s intention. We use the following training pipeline $C(.,\theta)$  to show how the diverse and unpaired captions are generated:
\begin{equation}
\begin{aligned}
\underbrace{C((\mathcal{S}|\mathcal{I});(\theta_{\mathcal{I}},\theta_{\mathcal{S}}))}_{\rm{TextCap\ Model}}&=  \underbrace{E_{\mathcal{I}}((\mathcal{G}^{\mathcal{I}}|\mathcal{I});\theta_{\mathcal{I}})}_{\rm{Image\ Encoder}} \\&
\!\!\!\!\!\!\!\underbrace{U_{\mathcal{A}}((\mathcal{G}^\mathcal{S}|\mathcal{G}^{\mathcal{I}});\theta_{\mathcal{I}})}_{\rm{Unpaired\ Alignment}}\underbrace{D_{\mathcal{S}}((\mathcal{S}|\mathcal{G}^\mathcal{S});\theta_{\mathcal{S}})}_{\rm{Sentence\ Decoder}}
\end{aligned}
\end{equation}
 $E_{\mathcal{I}}(\cdot; \theta)$ is image encoder aims to model $N_k$  multimodal relational graphs (MRGs) $\mathcal{G}^{\mathcal{I}}=\{\mathcal{G}^{\mathcal{I}}_i|^{N_k}_{i=1}\}$ learned from image $\mathcal{I}$, then translate $\mathcal{G}^{\mathcal{I}}$ to appropriate sentence graph $\mathcal{G}^{\mathcal{S}}$ through unpaired cross-domain alignment $U_{\mathcal{A}}(\cdot; \theta)$, the aligned results can be decoded to diverse captions $\mathcal{S}=\{\mathcal{S}_i|_{i=1}^{N_k}\}$ by sentence decoder $D_{\mathcal{S}}(\cdot; \theta)$. Notice that $E_{\mathcal{I}}(\cdot; \theta)$ and $U_{A}(\cdot; \theta)$ share parameters $\theta_{\mathcal{I}}$  to learn feature alignment across domain, $D_{\mathcal{S}}(\cdot; \theta)$ is a pretrained sentence docoder from textual corpus, $\theta_{\mathcal{S}}$ are fixed as explicit supervised signals to learn $\theta_{\mathcal{I}}$ to generate unpaired captions. 


Fig.2 shows the proposed Multimodal relAtional Graph adversarIal  inferenCe (MAGIC) framework $C(.;\theta)$  consists of three main modules: Multimodal Relational Graph Encoder $E_{\mathcal{I}}(\cdot; \theta)$, Sentence Auto-Encoder $D_{\mathcal{S}}(\cdot; \theta)$  and Unpaired Caption Adversarial Inference $U_{A}(\cdot; \theta)$.  We use the TextCap task as an example to illus- trate our method.

\subsection{Multimodal Relational Graph Encoder}
This module aims to select the central objects dynamically as the central object-aware pool, then constructs their MRGs adaptively, which can represent the different fine-grained contents. Last, GCN model intra- and cross-relationships to learn MRG embeddings.

\subsubsection{Multimodal Feature Extractor.}
Given a text-based image $\mathcal{I}$, we firstly initialize it as its corresponding $N_{o}$ visual object features and $N_{t}$ text token features by Faster RCNN~\cite{ren2016faster} and Rosetta OCR~\cite{borisyuk2018rosetta}, respectively. In order to represent visual and text features at a unified level, follow~\cite{sidorov2020textcaps}, we  adopt a single-layer MLP to map the above features as $d$-dimensional vectors, the unified visual and text  embedding sets  of image $\mathcal{I}$ given by $\textbf{O}=\{\textbf{o}_i|_{i=1}^{N_o} \}$ and $\textbf{T}=\{\textbf{t}_j|_{j=1}^{N_t}\}$, respectively. 




\subsubsection{Central Object-Aware Pool.}
Inspired by the human-being attention mechanism to describe an image, which usually focuses on a salient object or region, then from ``one-to-many'', queries its relevant information and ignores the unconcerned contents to compose fine-grained sentences. 
We imitate this process by introducing a central object-aware pool (COP) that  is supposed to reflect the human preference and select objects that may need additional attention adaptively. In detail, given 
$i^{th}$ object embedding $\textbf{o}_i$, we employ soft attention to  each text token embedding in $\textbf{T}$. Then  $\textbf{o}_i$ and attened textual latent vector $\textbf{t}_i^a$ are concatenated as the augmented object embedding $\textbf{o}_i^a=[\textbf{o}_i,\textbf{t}_i^a]$, the embedding set is given by $\textbf{O}^a=\{\textbf{o}_i^a|_{i=1}^{N_o}\}$. To decide the central objects from $\textbf{O}^a$, we adopt a MLP to predict a score for each object:
\begin{equation}
S^{c}_{object} =  {\rm softmax} (l_m^c(\textbf{O}^a))
\end{equation}
$S^{c}_{object}$ contains $N_o$ scores for each element that indicates the importance weight of each object. Then we select $N_k$ objects $\textbf{O}^c=\{\textbf{o}^c_i|_{i=1}^{N_k}\}$ with the highest scores as COP .


\subsubsection{Multimodal Relational Graph Embedding.} 
In analogy to the human inference of contextual information with a given focused object, we first query its intra- and cross- relationships for each central object, then construct their corresponding MRGs $\mathcal{G}^{\mathcal{I}}=\{\mathcal{G}^{\mathcal{I}}_i|^{N_k}_{i=1}\}$ to represent the diversity of image $\mathcal{I}$. In detail, for $k^{th}$ central object $\textbf{o}^c_k$,  its MRG defined as $\mathcal{G}_k^\mathcal{I} = (\mathcal{N}^\mathcal{I}, \mathcal{E}^\mathcal{I})$. We devote $\mathcal{N}^\mathcal{I}$ as three relational types according to the contextual information:  relationship nodes $\textbf{n}^r$  attribute nodes $\textbf{n}^a$ and text nodes $\textbf{n}^t$. On the one hand, intra-relational nodes $\textbf{n}^r$ and $\textbf{n}^a$ corresponds to the visually neighbour objects has relationships, attributes for object $\textbf{o}^c_k$, respectively. 
\begin{equation} \!\!\mathcal{N}^{\mathcal{I}}\{\textbf{n}^r, \textbf{n}^a\} \!\!= \!\!
\begin{cases} 
\{W_r \textbf{o}_i\}_{i=1}^{N_o-1}, \!\!\!& \mbox{relationship nodes};\\
\{W_a [\textbf{o}_i,s_i]\}_{i=1}^{N_o-1},  \!\!\!& \mbox{attribute nodes}. 
\end{cases} 
\end{equation}
where $W_r$  $\in \mathbb{R}^{ d \times d }$ and $W_a$  $\in \mathbb{R}^{ d \times (d+4) }$ are the relational embedding matrix, we omit the bias term for simplicity. $\textbf{s}_{i}$ the positional embedding to distinguish different attribute nodes connected with the object $\textbf{o}^c_k$.

On the other hand, to capture the cross-relationship among central object $\textbf{o}^c_k$ and  text tokens,  we perform soft attention weight on  texts $\textbf{T}$ to query the semantically textual embeddings as  text nodes $\textbf{n}^t$.
\begin{equation} 
\mathcal{N}^{\mathcal{I}}\{\textbf{n}^o\} = 
\{a_i^t \textbf{t}_i\}_{i=1}^{N_t-sum(a_i^t > \epsilon)}, \quad \mbox{text nodes}.
\end{equation}
where $a_i^t$ is the attention score, $\epsilon$ is a pre-defined threshold. $sum()$ is counting function aims to connect  more related contextual texts for  $o^c_k$ if attention score $a_j^t$ greater than $\epsilon$.  

Thus far, the MRG $\mathcal{G}_k^\mathcal{I} = (\mathcal{N}^\mathcal{I}, \mathcal{E}^\mathcal{I})$ with three relational nodes $\{\textbf{n}^r, \textbf{n}^a, \textbf{n}^t\}$ is obtained.
To model the semantic relations among the above MRG, we introduce a multi-layer GCN~\cite{marcheggiani2017encoding,guo2021semi}, which can capture
mutual relations between neighboured nodes and embed the graph structure into latent vectors.
\begin{equation} 
\textbf{g}^{\mathcal{I}}_k=  \delta_{r} (W_{m_0} \textbf{o}^c_k+ \sum\limits_{\hat{\textbf{n}}_i \in \mathcal{N}^{\mathcal{I}}} W_{m_1} \hat{\textbf{n}}_i)
\end{equation}
where  $\delta_{r}$ is the ReLU activation function, $W_{m_0}$ and $W_{m_1}$ $\in \mathbb{R}^{ d \times d }$ are the transformation matrices. In this multi-layer GCN, first layer brings contexts for each node from its direct neighbor nodes. Meanwhile, stacking multiple layers enables the encoding of broad contexts in the graph. After such graph pooling, the MRG embedding $\textbf{g}^{\mathcal{I}}_k$ for object $\textbf{o}^c_k$ is obtained.  As a consequence, the content diversity of image $\mathcal{I}$ can be regarded as the collection $\textbf{g}^{\mathcal{I}}=\{\textbf{g}_k^\mathcal{I}|_{k=1}^{N_k}\}$ of $N_k$ central objects.


\subsection{Sentence Auto-Encoder}
We design a sentence auto-encoder that fully capture the language properties by self-reconstructing sentence. In particular, we extract the scene graph $\mathcal{G}^{\mathcal{S}}$ from the sentence corpus, and then decode the sentence from latent vector that is encoded from the $\mathcal{G}^{\mathcal{S}}$ (sentence $\rightarrow$ $\mathcal{G}^{\mathcal{S}}$  $\rightarrow$ latent vector $g^{\mathcal{S}}$ $\rightarrow$ reconstructed sentence).

\subsubsection{Scene Graph Encoder.}
We use the scene graph parser provided by~\cite{anderson2016spice} for sentences, where a syntactic dependency tree is built by~\cite{klein2003accurate} and then a rule-based method~\cite{schuster2015generating} is applied for transforming the tree to a scene graph. 

Given a sentence $\hat{S}$ from TextCap corpus, its corresponding scene graph involved the high-level semantic concepts (e.g., objects, relationships  between two adjacent objects and attributes of objects) and textual tokens, to align subsequently image-sentence domain better, we encode sentences in a similar way in MRG encoding.
In detail, for each conceptual object $\textbf{c}_i^o$ in $\hat{S}$, we define its relational graph as $\mathcal{G}_i^{\mathcal{S}}=(\mathcal{N}^\mathcal{S}, \mathcal{E}^\mathcal{S})$, where $\mathcal{N}^s$ devote three relational type, relationship nodes $\textbf{c}^r$, attribute nodes $\textbf{c}^a$ and associated textual nodes $\textbf{c}^t$. The global semantic meaning of sentence $\hat{S}$  can be regarded as the collection of object relational graphs $\mathcal{G}^{\mathcal{S}}=\{\mathcal{G}^{\mathcal{S}}_i|^{M_o}_{i=1}\}$, where $M_{o}$ is the object number in sentence $\mathcal{S}$. Similarly, the latent embedding of sentence is learned by the multi-layer GCN which using same structure in image graph encoding but independent parameters.
\begin{equation} 
g^{\mathcal{S}}=  \frac{1}{M_{o}}\delta_{r} ( \sum\limits_{i=1 }^{M_o}( W_{c0} \textbf{c}^o_i+ \sum\limits_{\hat{\textbf{c}}_i \in \mathcal{N}^{\mathcal{S}} } W_{c1} \hat{\textbf{c}}_i))
\end{equation}
where $W_{c0}$ and $W_{c1}$ $\in \mathbb{R}^{ d \times e }$,   Thus, the  scene graph embedding of sentence $\mathcal{\hat{S}}$ is given by $\textbf{g}^{\mathcal{S}}$.

\subsubsection{Sentence Decoder.}
The decoder part is essential in the way that it translates the semantic latent vector $\textbf{g}^{\mathcal{S}}$ into the textual sentence. However, most text tokens in sentence corpus only appear few times. This long-tailed word distribution specifically demonstrates the large variance in text occurring, which is challenging to model using a fixed word vocabulary. For this challenge, following M4C~\cite{sidorov2020textcaps}, we adopt different classifiers for common vocabulary and candidate texts to predict words. In detail, to capture long-range dependency in sequence modeling, a general LSTM is employed to predict the word $\hat{\textbf{y}}_{t}$ in common vocabulary based on the scene graph embedding
$\textbf{g}^{\mathcal{S}}$. Then a dynamic pointer network that makes final word prediction on the basis of text candidates $\textbf{T}^c$ and $\hat{\textbf{y}}_{t}$.
\begin{equation} 
\begin{aligned}
\hat{\textbf{y}}_{t} &= l_m^w({\rm LSTM_{DEC}} (\textbf{h}_{t-1}, \textbf{g}^{\mathcal{S}})),\\
\textbf{y}^t&= {\rm argmax}([\hat{\textbf{y}}_{t}, DPN(\hat{\textbf{y}}_{t},\textbf{T}^c))]);
\end{aligned}
\end{equation}
where $l_m^w(.)$ is a linear classifier for common vocabulary, $DPN(.)$ denotes the dynamic pointer network. We devote notation $\textbf{y}_{0:T}$ refers to a sequence of predicted words $\{\textbf{y}_0, \cdots, \textbf{y}_T \}$,  the captioning loss can be computed by:

\begin{equation}\label{10}
\mathcal{L}(\mathcal{S};\theta_{\mathcal{S}}) =  -\rm{log} \cdot \sum^T_{i=0} (P(\textbf{y}_t| (\textbf{y}_{0:T-1}, \textbf{g}^{\mathcal{S}}, \textbf{T}^c)))
\end{equation}


\subsection{Unpaired Adversarial Caption Inference}
To achieve this unpaired captioning, we develop a CGAN that contains two modules, namely, cross-domain alignment and language discriminator. We first perform the feature alignment across domains and then distinguish whether the semantic structure of decoded caption is real or fake using the language discriminator. Such a cascaded mechanism can train a captioning model in an unsupervised manner and improve coherence among generated sentences.
\subsubsection{Cross-Domain Alignment.}
Given two unpaired graph embeddings sets $\textbf{G}^{\mathcal{I}}=\{\textbf{g}_i^\mathcal{I}|_{i=1}^{N_{\mathcal{I}}}\}$ and $\textbf{G}^{\mathcal{S}}=\{\textbf{g}_i^\mathcal{S}|_{i=1}^{N_{\mathcal{S}}}\}$ from image and sentence domain, respectively, where $N_{\mathcal{I}}$ and $N_{\mathcal{S}}$ are the capacity of corresponding corpus.  The goal of feature alignment is enforce  the $\textbf{G}^{\mathcal{I}}$  to be close to the $\textbf{G}^{\mathcal{S}}$  distribution. We follow ~\cite{zhu2017unpaired,gu2019unpaired} that align feature space through cycle-consistent adversarial training. Differently, considering that the huge domain discrepancy, we leverage recent advances in Optimal Transport~\cite{peyre2019computational} to encourage the cross-modal semantic coherence.  In detail, the  wasserstein distance~\cite{shen2018wasserstein} is employed as the loss value to learn the discriminator, to align domain $X$ to domain $Y$, the distribution distance $W(X, Y)$ can be defined and corresponding loss can be minimized as follows:
 \begin{equation}
\begin{aligned}
W(X, Y) &= \inf_{\gamma \sim \Pi (X, Y)} \mathbb{E}_{(\textbf{x}, \textbf{y}) \sim\gamma} [||\textbf{x}- \textbf{y}||]\\
\mathcal{L}({X \!\!\rightarrow\!\! Y};\theta)&= \mathbb{E}_{\textbf{y} \sim Y}[D_{Y}(\textbf{y})]-\mathbb{E}_{\textbf{x} \sim X}[D_{Y}(\textbf{x})]
\end{aligned}
\end{equation}
 where $\gamma \sim \Pi (X, Y)$ is the set of possible joint distributions for the combination of $\textbf{x}$ and $\textbf{y}$. $D_{Y}$ is a discriminator to distinguish the origin source of the latent vector if from  $Y$.  Such loss is more effectively maintain the semantic coherence between two domain with huge discrepant characteristics. Thus, the cycle loss of feature alignment from $\textbf{G}^{\mathcal{I}}$ to $\textbf{G}^{\mathcal{S}}$ can be defined as follows:
 \begin{equation}
 \begin{aligned}
\mathcal{L}({A};\theta_{\mathcal{I}})=&\lambda_A (\mathcal{L}(M_\mathcal{S}(\textbf{G}^{\mathcal{I}}) \!\! \rightarrow \!\! \textbf{G}^{\mathcal{S}})+\mathcal{L}(M_\mathcal{I}(\textbf{G}^{\mathcal{S}})\\& \!\! \rightarrow \!\! \textbf{G}^{\mathcal{I}})+\lambda_C\mathcal{L}(M_\mathcal{I}(\textbf{G}^{\mathcal{I}})\!\! \leftrightarrow  \!\! M_\mathcal{I}(\textbf{G}^{\mathcal{S}})))
\end{aligned}
\end{equation}
where $\lambda_A$ and $\lambda_C$ are weight hyper-parameters to control the feature alignment, $M_\mathcal{I}$ and $M_\mathcal{S}$ are mapping function to image domain and sentence domain, respectively. $\mathcal{L}(M_\mathcal{I}(\textbf{G}^{\mathcal{I}})\!\! \leftrightarrow  \!\! M_\mathcal{I}(\textbf{G}^{\mathcal{S}}))$ is the  cycle consistency loss to regularize the training. Subsequently, the aligned image-sentence embeddings $\textbf{G}^{\mathcal{\tilde{S}}}=\{\tilde{\textbf{{g}}}_i^\mathcal{S}|_{i=1}^{N_{\mathcal{I}}}\}$ and text candidates $T^c$ are feed into sentence decoder to generate captions.
\subsubsection{Language Discriminator.} 

Notably, the cycle loss can minimize the distribution distance among image and sentence domains. However, the characteristic distribution of aligned image embeddings  $\textbf{G}^{\mathcal{\tilde{S}}}=\{\tilde{\textbf{{g}}}_i^\mathcal{S}|_{i=1}^{N_{\mathcal{I}}}\}$ is between image domain and sentence domain. To capture the language characteristics further, we introduce a language discriminator $D_L$ that focuses on the language structure of an individual sentence. Here, we want to ensure fluency and language accuracy that may be lacking in the cross-domain alignment. In addition to generated captions from the sentence decoder,  $D_L$ is given negative inputs with a mixture of randomly shuffled words or repeated phrases within a sentence.
\begin{equation}
\begin{aligned}
\mathcal{L}(L;\theta_{\mathcal{S}})=&  \lambda_L( \mathbb{E}_{\textbf{s} \sim S}[\rm{log}(D_{ {L}}(\textbf{s}))]- \mathbb{E}_{\textbf{s}^r \sim S^r}\\&[1-  \rm{log}(D_{ {L}}(\textbf{s}^r))])
 \end{aligned}
\end{equation}
where $\lambda_L$ is weight hyper-parameter and $D_L$ is a Bidirectional LSTM. $S^r$ is a set in which sentences with random N-grams. This language discriminator verifies not only the semantic fluency among sentences but also improves the captioning accuracy.

\begin{table*}[t]
  \centering
    \begin{tabular}{l |c| c c c c c| c c c c}
    \hline
     &\multicolumn{1}{|c|}{DIC.} &\multicolumn{5}{c|}{ Captioning Evaluation } &\multicolumn{4}{c}{ Diversity Evaluation }\\
    \hline
    Method &  & BLEU & METEOR  & ROUG  &  SPICE &   CIDEr& Div-1  & Div-2   & RE-4 $\downarrow$& SelfCIDEr \\
    \hline
     \hline
    ISM-LSTM      &   & 15.2 &17.6& 38.4 &11.8 &31.1 &25.2 &34.2& 5.3 &41.9\\
    \hline
    MAGIC (w/o CO)$^{\dagger}$      &   & 20.2 &18.1& 39.5& 12.3 &71.3 &28.8 &37.4& 4.8 &45.4\\
     MAGIC (w/o TA)$^{\dagger}$    &\checkmark & \textbf{21.7}&	\textbf{20.3}&	\textbf{42.0}&	\textbf{13.6}&	\textbf{75.9} &\textbf{28.1} &\textbf{40.4}& \textbf{3.6} &\textbf{49.1}\\
    MAGIC (w/o CG)$^{\dagger}$       &\checkmark & 20.1 &18.4& 37.1& 12.2 &65.3 &26.8 &36.2& 5.3 &46.1\\
    MAGIC (w/o LD)$^{\dagger}$                               &\checkmark & 20.8 &20.1& 41.1& 12.8 &73.9 &28.2 &38.3& 4.5 &48.8\\
    \hline
     MAGIC                                          &\checkmark & \textbf{22.2}&	\textbf{20.5}&	\textbf{42.3}&	\textbf{13.8}&	\textbf{76.6} &\textbf{29.7} &\textbf{40.9}& \textbf{3.8} &\textbf{49.6}\\
       
    \hline
    \end{tabular}
    \caption{Captioning and diversity results on TextCap validation set. DIC. indicates the results from diverse image captions.   $^{\dagger}$ indicates whether the model is a variant of MAGIC. Larger values indicate better performance, except for small value is better for RE-4. The top two scores on each metric are in bold. Acronym notations of each method see in comparison of methods.}
    \vspace{-0.2cm}
\end{table*}
\begin{table}[t]
  \centering
    \begin{tabular}{l |c c c c  }
    \hline
     & \multicolumn{4}{c}{ Results of differnet COP Size}\\
    \hline
    COP Capacity  &BLEU& CIDEr & Div-2 & SelfCIDEr  \\
    \hline
    \hline
    $N_k$=1      &\textbf{22.3}& \textbf{77.3} & \textbf{41.9}& \textbf{50.3} \\
   $N_k$=2      &21.8& {75.8} & \textbf{41.1}& {49.3} \\
      $N_k$=4     &21.6& 75.3 & 39.3& 47.8\\
    $N_k$=5     & 20.8& 73.4 & 38.1& 47.5\\
    \hline
    $N_k$=3    & \textbf{22.2}& \textbf{76.6} & 40.9& \textbf{49.6} \\
    \hline
    \end{tabular}
    \caption{ The captioning and diversity results of different sike $N_k$ of central object-aware pool. }
    \vspace{-0.1cm}
\end{table}
\subsection{Training Algorithm}
Algorithm 1 (in the appendix) presents the pseudocode of our MAGIC algorithm for diverse and unpaired TextCap. First, we pretrain the sentence auto-encoder by reconstructing sentences and the language discriminator as well as disrupting the order of words from the sentence corpus. The MAGIC framework then learns diverse multimodal relational graph embeddings by querying the semantic meaning of central objects. Finally, the developed CGAN aligns cross-modal embeddings from the image domain to the sentence domain and then checks the fluency and accuracy of generated captions using the language discriminator to generate unpaired and diverse captions magically.
\section{Experiments}
We first benchmark our MAGIC for unpaired and diverse captioning on the TextCaps dataset~\cite{sidorov2020textcaps}  to verify its effectiveness, and then discuss MAGIC’s property with controlled studies.
\subsection{Dataset and Setting}
\noindent$\textbf{Dataset.}$ The TextCaps dataset collected 28,408 images with  142,040 captions from Open Image V3 dataset. Each image has five captions with verified texts through the Rosetta OCR system~\cite{borisyuk2018rosetta} and human annotators.  Follows TextVQA ~\cite{singh2019towards}, this dataset splits for training (21,953), validation (3,166), and test (3,289) sets.


\noindent$\textbf{Implementation Details.}$  The number of visual objects is $N_o$ = 100 and the number of OCR tokens is $N_t$ = 50. For TextCap generation, we tokenized the texts on whitespace, and the sentences are ``cut'' at a maximum length of 20 words. We add a \emph{Unknown token} to replace the words out of the vocabulary list. The vocabulary has 6,736 words, and each word is represented by a 300-dimensional vector.



\noindent \textbf{Metrics.} We use the microsoft coco caption evaluation \footnote{https://github.com/tylin/coco-caption/}, which includes BLEU~\cite{papineni2002bleu} , METEOR~\cite{denkowski2014meteor}, ROUG\_L~\cite{rouge2004package},  CIDEr~\cite{vedantam2015cider}  and SPICE~\cite{anderson2016spice} to evaluate the captioning results.  To evaluate the captioning diversity, we use Div-1 and Div-2 scores~\cite{li2015diversity} that measure a ratio of unique N-grams (N=1,2) to the total number of words. We also report RE-4 that captures a degree of N-gram repetition (N=4) in a description and SelfCIDEr~\cite{wang2019describing} to study the semantic level diversity.

\noindent \textbf{Comparison of Methods.}  Diverse and unpaired Textcap is a fire-new task that existing captioning algorithms are not applied to compare. We simplified MAGIC as the baseline and conducted an ablation study to ensure a fair comparison and investigate contributions of individual components in MAGIC using the following versions: \textbf{ISM-LSTM} is a simplified version of MAGIC that maps image features from the image corpus to the sentence corpus using cycle GAN and then decodes the caption with LSTM; \textbf{MAGIC (w/o CO)} is similar to~\cite{yao2018exploring} and encodes the global spatial graph as the image embedding to decode a single caption; \textbf{MAGIC (w/o TA)} considers all text tokens in the MRGs without query attention to examine the impact on the diversity of captions; \textbf{MAGIC (w/o CG)} utilizes the original gan to align the image–sentence domain to investigate the impact on the accuracy of captions.; ~\textbf{MAGIC (w/o LD)} generates captions without the language discriminator to assess the importance of the cascaded adversarial mechanism.

\subsection{Experimental Results}
\noindent$\textbf{Captioning Results.}$ The overall quantitative results of our model and baseline on the validation set of the Textcap dataset are listed in Tab.~1. The trained baseline model (ISM-LSTM) achieves the minimum captioning score, thereby indicating that it fails to describe the text-based image due to missing text awareness. The excellent performance of MAGIC and its variants in comprehending multimodal content compared with the baseline verifies the effectiveness of our proposed model. Notably, the results of other variants are better than those of MAGIC (w/o CO) and MAGIC (w/o CG). Our analysis showed that MAGIC (w/o CO) encodes an image with global content with abundant multimodal details, but its ground truth captions describe the image locally. The CGAN may fail to align the cross-domain appropriately. Compared with MAGIC, MAGIC (w/o CO) with original GAN may fail to capture cross-domain semantic consistency effectively. In contrast, MAGIC with CGAN can improve the quality of unpaired captions in feature alignment and semantic structure levels.
\begin{table}[t]
  \centering
    \begin{tabular}{l |c c c c  }
    \hline
     & \multicolumn{4}{c}{ Results of different COP construction}\\
    \hline
    Central Object  &BLEU& CIDEr & Div-2 & SelfCIDEr  \\
    \hline
    \hline
   Center      & {18.7}& 72.3 & {38.3}& {44.2} \\
     Large    &  \textbf{20.1}& \textbf{75.9} & \textbf{40.2}& \textbf{49.3}\\
      Random     &  {17.1}& 67.8 & {36.9}& {46.6}\\
     \hline
  Top Socre     & \textbf{22.2}& \textbf{76.6} & \textbf{40.9}& \textbf{49.6} \\
    \hline
    \end{tabular}
    \caption{ The captioning and diversity results of different COP construction. Center, large and random means that the central object is selected by different rules.}
    \vspace{-0.2cm}
\end{table} 
\begin{figure*}[t]
\includegraphics[width=1\textwidth]{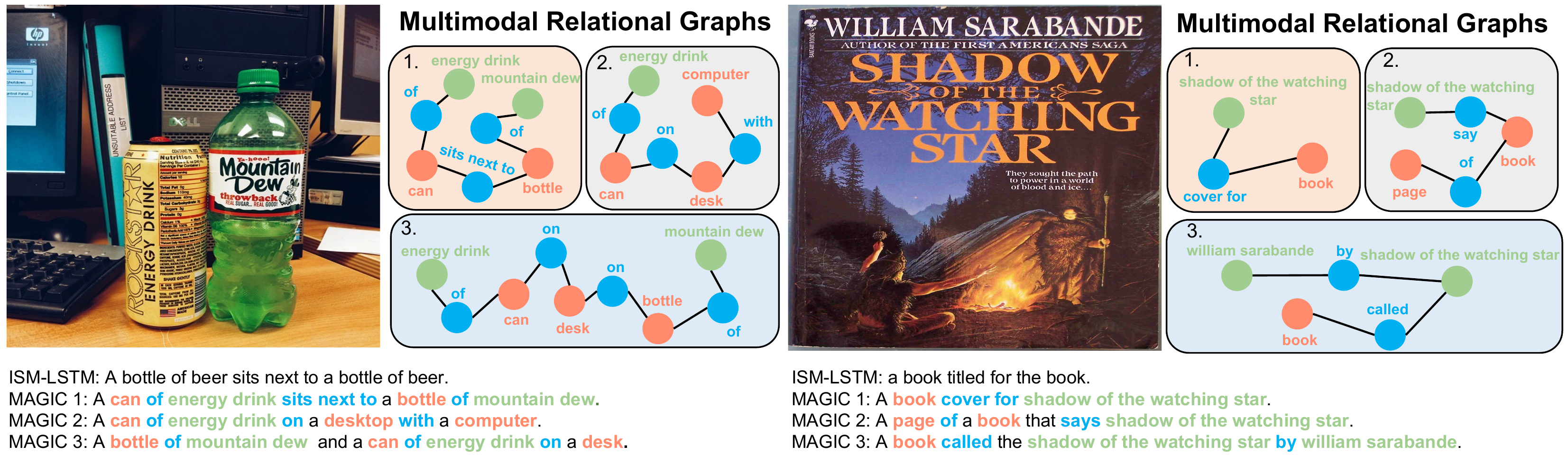}
\centering\caption{Two qualitative cases of MAGIC from TextCaps validation set. For better visualisation, we contruct the multimodal relational graphs based on generated diverse captions. }
\vspace{-0.2cm}
\end{figure*}

\begin{figure}[t]
\includegraphics[width=0.5\textwidth]{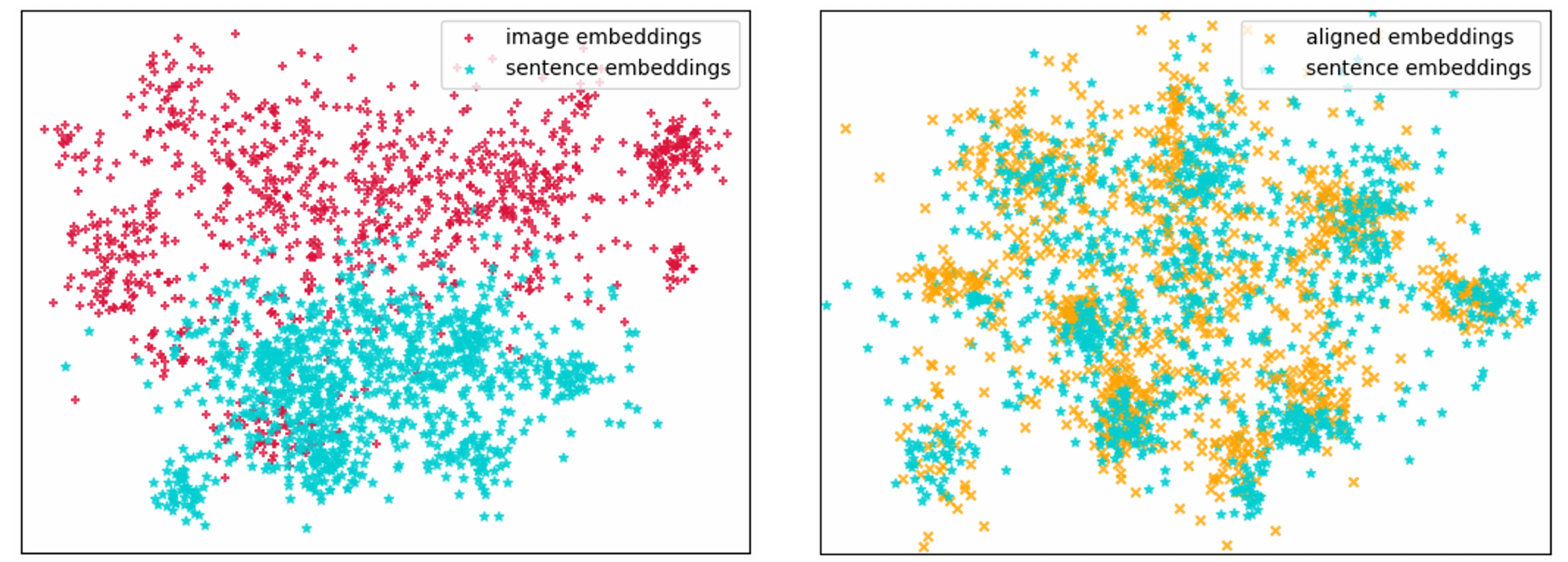}
\centering\caption{Visualization of features (1000 samples) 2D space by t-SNE~\cite{van2008visualizing}. }
\vspace{-0.1cm}
\end{figure}
\noindent$\textbf{Diversity Results.}$ Tab.~1 also presents the comparison of diversity results. The top three objects are selected as the central object-aware pool in MAGIC variants to ensure a fair comparison. Notably, diverse captioning models of MAGIC outperform the single captioning baseline of ISM-LSTM by a large margin. Generated various captions effectively respect MRGs, especially in the SelfCider score, which focuses on semantic similarity. We conduct additional diversity analyses in the following section.

\subsection{In-Depth Analysis}
We further validate the several key issues of the proposed method by answering three questions as follows. 

$\textbf{Q1: How does the structure of central object-aware}$\\
$\textbf{pool (COP) affect the captioning diversity?}$ On the one hand, we conduct an ablation study to illustrate the impact of different sizes of COP  on diverse and unpaired TextCap. 
 The results are presented in Tab.~2. MAGIC with a pool size of $N_k$=1,2,3 presents the solid and approximate TextCap results, and MAGIC with a pool size of $N_k$=4,5 obtains competitive results. For this situation,  we found that there are objective semantic overlaps among annotated image descriptions, i.e., for an image, some ground truths of an image from the sentence corpus are similarly semantic in the sentence domain.  This similarity may cause confusion in the alignment of different multimodal relational graphs to the sentence domain in CGAN and lead to slight performance degradation. On the other hand, the quantitative results according to rule- and learning-based central object selections for OAP are reported. Tab.3 demonstrated that all rule-based methods suffer from low metric performance while the learning-based central object can accurately and diversely describe an image.

$\textbf{Q2: How much improvement of unpaired TextCap}$\\
$\textbf{quality has the MAGIC brought?}$ 
The MAGIC can improve the unpaired TextCap effectively by feature alignment and semantic coherence distinguishing.  First, we visualize the feature alignment to indicate that our MAGIC can capture the semantic consistency across the image–sentence domain Fig.~4. 
To better understand how satisfactory are the sentences generated by CGAN, we  conducted a human study with five experienced workers to evaluate the generated descriptions from MAGIC and MAGIC (w/o CG). Workers evaluated 100 images which are randomly sampled from the validation set for each pairwise comparison.  Fig.~5 shows that employing MAGIC can describe images more diversely, accurately and fluently.  

\begin{figure}[t]
\vspace{-0.1cm}
\includegraphics[width=0.5\textwidth]{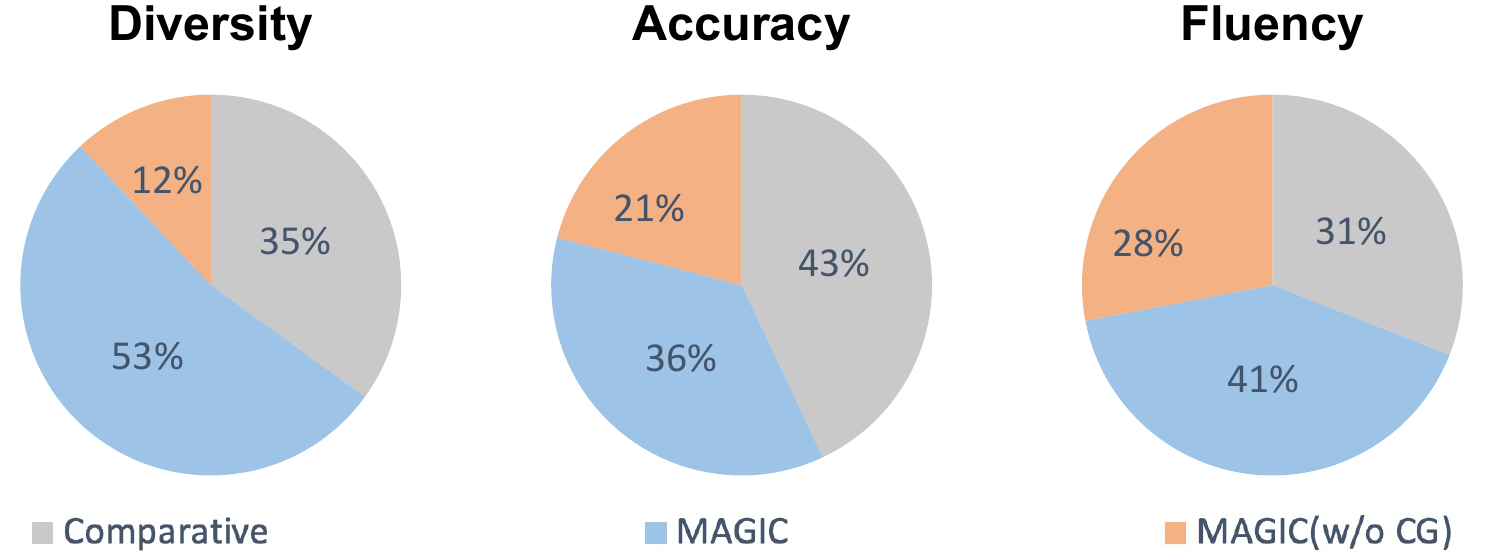}
\centering\caption{The pie charåts each comparing the two methods in human evaluation.  }
\vspace{-0.2cm}
\end{figure}
$\textbf{Q3: How about the diversity of the visual concepts and}$\\$\textbf{texts for an image with multi-captions?}$ We additionally conduct a qualitative analysis of the descriptive diversity of MAGIC in Fig.~3. We present the multiple MRGs ($N_k$=3) for each image on the basis of generated captions. The excellent results showed that MAGIC produces more accurate and fine-grained captions than baseline. In this case(a), MAGIC models complex multimodal relationships correctly and describes an image with different fine-grained contents adaptively. As presented in third caption, for intra-relationship, MAGIC can capture the semantic relationship ``\emph{can-sit nexts to-bottle}", ``\emph{bottle-on-desk}", and ``\emph{can-on-desk}" correctly; for cross-relationship, MAGIC can model the semantic relationship  ``\emph{bottle-of-energy drink}" and ``\emph{can-of-mountain dew}" appropriately. However, the baseline ISM-LSTM describes the image with error semantic information and semantic structure.

\section{Conclusion}
We focus on diverse and unpaired text-based image caption generation. The MAGIC framework is proposed to achieve these objectives. MAGIC can adaptively build multiple multimodal relational graphs on the basis of selected central objects. A cascaded generative adversarial network is then developed to infer the unpaired caption generation in the image–sentence feature alignment level and linguistic coherence. The experimental results verified that MAGIC can effectively generate very promising and diverse captions without using any paired training data. The results of our study can provide new insights into visual captioning and guidance for future investigations on vision and language.

\section{ Acknowledgments} This work has been supported in Apart by National Key Research and Development Program of China (2018AAA0101900).

\bibliographystyle{aaai21}
\bibliography{aaai22}

\begin{thebibliography}{45}
\providecommand{\natexlab}[1]{#1}

\bibitem[{Anderson et~al.(2016)Anderson, Fernando, Johnson, and
  Gould}]{anderson2016spice}
Anderson, P.; Fernando, B.; Johnson, M.; and Gould, S. 2016.
\newblock Spice: Semantic propositional image caption evaluation.
\newblock In \emph{European conference on computer vision}, 382--398. Springer.

\bibitem[{Berg, Kipf, and Welling(2017)}]{berg2017graph}
Berg, R. v.~d.; Kipf, T.~N.; and Welling, M. 2017.
\newblock Graph convolutional matrix completion.
\newblock \emph{arXiv preprint arXiv:1706.02263}.

\bibitem[{Borisyuk, Gordo, and Sivakumar(2018)}]{borisyuk2018rosetta}
Borisyuk, F.; Gordo, A.; and Sivakumar, V. 2018.
\newblock Rosetta: Large scale system for text detection and recognition in
  images.
\newblock In \emph{Proceedings of the 24th ACM SIGKDD International Conference
  on Knowledge Discovery \& Data Mining}, 71--79.

\bibitem[{Caron et~al.(2020)Caron, Misra, Mairal, Goyal, Bojanowski, and
  Joulin}]{caron2020unsupervised}
Caron, M.; Misra, I.; Mairal, J.; Goyal, P.; Bojanowski, P.; and Joulin, A.
  2020.
\newblock Unsupervised learning of visual features by contrasting cluster
  assignments.
\newblock \emph{arXiv preprint arXiv:2006.09882}.

\bibitem[{Chen et~al.(2021)Chen, Pu, Wu, Xie, Liu, and Lin}]{chen2021cross}
Chen, T.; Pu, T.; Wu, H.; Xie, Y.; Liu, L.; and Lin, L. 2021.
\newblock Cross-domain facial expression recognition: A unified evaluation
  benchmark and adversarial graph learning.
\newblock \emph{IEEE transactions on pattern analysis and machine
  intelligence}.

\bibitem[{Cornia et~al.(2020)Cornia, Stefanini, Baraldi, and
  Cucchiara}]{cornia2020meshed}
Cornia, M.; Stefanini, M.; Baraldi, L.; and Cucchiara, R. 2020.
\newblock Meshed-memory transformer for image captioning.
\newblock In \emph{Proceedings of the IEEE/CVF Conference on Computer Vision
  and Pattern Recognition}, 10578--10587.

\bibitem[{Denkowski and Lavie(2014)}]{denkowski2014meteor}
Denkowski, M.; and Lavie, A. 2014.
\newblock Meteor universal: Language specific translation evaluation for any
  target language.
\newblock In \emph{Proceedings of the ninth workshop on statistical machine
  translation}, 376--380.

\bibitem[{Ding et~al.(2020)Ding, Qu, Xi, and Wan}]{ding2020stimulus}
Ding, S.; Qu, S.; Xi, Y.; and Wan, S. 2020.
\newblock Stimulus-driven and concept-driven analysis for image caption
  generation.
\newblock \emph{Neurocomputing}, 398: 520--530.

\bibitem[{Gu et~al.(2018)Gu, Joty, Cai, and Wang}]{gu2018unpaired}
Gu, J.; Joty, S.; Cai, J.; and Wang, G. 2018.
\newblock Unpaired image captioning by language pivoting.
\newblock In \emph{Proceedings of the European Conference on Computer Vision
  (ECCV)}, 503--519.

\bibitem[{Gu et~al.(2019)Gu, Joty, Cai, Zhao, Yang, and Wang}]{gu2019unpaired}
Gu, J.; Joty, S.; Cai, J.; Zhao, H.; Yang, X.; and Wang, G. 2019.
\newblock Unpaired image captioning via scene graph alignments.
\newblock In \emph{Proceedings of the IEEE/CVF International Conference on
  Computer Vision}, 10323--10332.

\bibitem[{Guo et~al.(2021)Guo, Shi, Kang, Kuang, Tang, Jiang, Sun, Wu, and
  Zhuang}]{guo2021semi}
Guo, J.; Shi, H.; Kang, Y.; Kuang, K.; Tang, S.; Jiang, Z.; Sun, C.; Wu, F.;
  and Zhuang, Y. 2021.
\newblock Semi-Supervised Active Learning for Semi-Supervised Models: Exploit
  Adversarial Examples With Graph-Based Virtual Labels.
\newblock In \emph{Proceedings of the IEEE/CVF International Conference on
  Computer Vision}, 2896--2905.

\bibitem[{Guo et~al.(2019)Guo, Liu, Yao, Li, and Lu}]{guo2019mscap}
Guo, L.; Liu, J.; Yao, P.; Li, J.; and Lu, H. 2019.
\newblock Mscap: Multi-style image captioning with unpaired stylized text.
\newblock In \emph{Proceedings of the IEEE/CVF Conference on Computer Vision
  and Pattern Recognition}, 4204--4213.

\bibitem[{Hou et~al.(2019)Hou, Wu, Qi, Zhao, Luo, and Jia}]{hou2019relational}
Hou, J.; Wu, X.; Qi, Y.; Zhao, W.; Luo, J.; and Jia, Y. 2019.
\newblock Relational reasoning using prior knowledge for visual captioning.
\newblock \emph{arXiv preprint arXiv:1906.01290}.

\bibitem[{Johnson, Karpathy, and Fei-Fei(2016)}]{johnson2016densecap}
Johnson, J.; Karpathy, A.; and Fei-Fei, L. 2016.
\newblock Densecap: Fully convolutional localization networks for dense
  captioning.
\newblock In \emph{Proceedings of the IEEE conference on computer vision and
  pattern recognition}, 4565--4574.

\bibitem[{Klein and Manning(2003)}]{klein2003accurate}
Klein, D.; and Manning, C.~D. 2003.
\newblock Accurate unlexicalized parsing.
\newblock In \emph{Proceedings of the 41st annual meeting of the association
  for computational linguistics}, 423--430.

\bibitem[{Lample et~al.(2017)Lample, Conneau, Denoyer, and
  Ranzato}]{lample2017unsupervised}
Lample, G.; Conneau, A.; Denoyer, L.; and Ranzato, M. 2017.
\newblock Unsupervised machine translation using monolingual corpora only.
\newblock \emph{arXiv preprint arXiv:1711.00043}.

\bibitem[{Li et~al.(2015)Li, Galley, Brockett, Gao, and
  Dolan}]{li2015diversity}
Li, J.; Galley, M.; Brockett, C.; Gao, J.; and Dolan, B. 2015.
\newblock A diversity-promoting objective function for neural conversation
  models.
\newblock \emph{arXiv preprint arXiv:1510.03055}.

\bibitem[{Li et~al.(2019)Li, Tang, Wu, and Zhuang}]{li2019walking}
Li, J.; Tang, S.; Wu, F.; and Zhuang, Y. 2019.
\newblock Walking with mind: Mental imagery enhanced embodied qa.
\newblock In \emph{Proceedings of the 27th ACM International Conference on
  Multimedia}, 1211--1219.

\bibitem[{Li et~al.(2021)Li, Tang, Zhu, Shi, Huang, Wu, Yang, and
  Zhuang}]{li2021adaptive}
Li, J.; Tang, S.; Zhu, L.; Shi, H.; Huang, X.; Wu, F.; Yang, Y.; and Zhuang, Y.
  2021.
\newblock Adaptive hierarchical graph reasoning with semantic coherence for
  video-and-language inference.
\newblock In \emph{Proceedings of the IEEE/CVF International Conference on
  Computer Vision}, 1867--1877.

\bibitem[{Li et~al.(2020)Li, Wang, Tang, Shi, Wu, Zhuang, and
  Wang}]{li2020unsupervised}
Li, J.; Wang, X.; Tang, S.; Shi, H.; Wu, F.; Zhuang, Y.; and Wang, W.~Y. 2020.
\newblock Unsupervised reinforcement learning of transferable meta-skills for
  embodied navigation.
\newblock In \emph{Proceedings of the IEEE/CVF Conference on Computer Vision
  and Pattern Recognition}, 12123--12132.

\bibitem[{Liu et~al.(2021)Liu, Gao, Zhang, and Zou}]{liu2021exploring}
Liu, F.; Gao, M.; Zhang, T.; and Zou, Y. 2021.
\newblock Exploring Semantic Relationships for Unpaired Image Captioning.
\newblock \emph{arXiv preprint arXiv:2106.10658}.

\bibitem[{Marcheggiani and Titov(2017)}]{marcheggiani2017encoding}
Marcheggiani, D.; and Titov, I. 2017.
\newblock Encoding sentences with graph convolutional networks for semantic
  role labeling.
\newblock \emph{arXiv preprint arXiv:1703.04826}.

\bibitem[{Papineni et~al.(2002)Papineni, Roukos, Ward, and
  Zhu}]{papineni2002bleu}
Papineni, K.; Roukos, S.; Ward, T.; and Zhu, W.-J. 2002.
\newblock Bleu: a method for automatic evaluation of machine translation.
\newblock In \emph{Proceedings of the 40th annual meeting of the Association
  for Computational Linguistics}, 311--318.

\bibitem[{Peyr{\'e}, Cuturi et~al.(2019)}]{peyre2019computational}
Peyr{\'e}, G.; Cuturi, M.; et~al. 2019.
\newblock Computational optimal transport: With applications to data science.
\newblock \emph{Foundations and Trends{\textregistered} in Machine Learning},
  11(5-6): 355--607.

\bibitem[{Ren et~al.(2016)Ren, He, Girshick, and Sun}]{ren2016faster}
Ren, S.; He, K.; Girshick, R.; and Sun, J. 2016.
\newblock Faster R-CNN: towards real-time object detection with region proposal
  networks.
\newblock \emph{IEEE transactions on pattern analysis and machine
  intelligence}, 39(6): 1137--1149.

\bibitem[{Rennie et~al.(2017)Rennie, Marcheret, Mroueh, Ross, and
  Goel}]{rennie2017self}
Rennie, S.~J.; Marcheret, E.; Mroueh, Y.; Ross, J.; and Goel, V. 2017.
\newblock Self-critical sequence training for image captioning.
\newblock In \emph{Proceedings of the IEEE Conference on Computer Vision and
  Pattern Recognition}, 7008--7024.

\bibitem[{ROUGE(2004)}]{rouge2004package}
ROUGE, L.~C. 2004.
\newblock A package for automatic evaluation of summaries.
\newblock In \emph{Proceedings of Workshop on Text Summarization of ACL,
  Spain}.

\bibitem[{Schuster et~al.(2015)Schuster, Krishna, Chang, Fei-Fei, and
  Manning}]{schuster2015generating}
Schuster, S.; Krishna, R.; Chang, A.; Fei-Fei, L.; and Manning, C.~D. 2015.
\newblock Generating semantically precise scene graphs from textual
  descriptions for improved image retrieval.
\newblock In \emph{Proceedings of the fourth workshop on vision and language},
  70--80.

\bibitem[{Shen et~al.(2018)Shen, Qu, Zhang, and Yu}]{shen2018wasserstein}
Shen, J.; Qu, Y.; Zhang, W.; and Yu, Y. 2018.
\newblock Wasserstein distance guided representation learning for domain
  adaptation.
\newblock In \emph{Thirty-Second AAAI Conference on Artificial Intelligence}.

\bibitem[{Sidorov et~al.(2020)Sidorov, Hu, Rohrbach, and
  Singh}]{sidorov2020textcaps}
Sidorov, O.; Hu, R.; Rohrbach, M.; and Singh, A. 2020.
\newblock Textcaps: a dataset for image captioning with reading comprehension.
\newblock In \emph{European Conference on Computer Vision}, 742--758. Springer.

\bibitem[{Singh et~al.(2019)Singh, Natarajan, Shah, Jiang, Chen, Batra, Parikh,
  and Rohrbach}]{singh2019towards}
Singh, A.; Natarajan, V.; Shah, M.; Jiang, Y.; Chen, X.; Batra, D.; Parikh, D.;
  and Rohrbach, M. 2019.
\newblock Towards vqa models that can read.
\newblock In \emph{Proceedings of the IEEE/CVF Conference on Computer Vision
  and Pattern Recognition}, 8317--8326.

\bibitem[{Van~der Maaten and Hinton(2008)}]{van2008visualizing}
Van~der Maaten, L.; and Hinton, G. 2008.
\newblock Visualizing data using t-SNE.
\newblock \emph{Journal of machine learning research}, 9(11).

\bibitem[{Vedantam, Lawrence~Zitnick, and Parikh(2015)}]{vedantam2015cider}
Vedantam, R.; Lawrence~Zitnick, C.; and Parikh, D. 2015.
\newblock Cider: Consensus-based image description evaluation.
\newblock In \emph{Proceedings of the IEEE conference on computer vision and
  pattern recognition}, 4566--4575.

\bibitem[{Wang and Chan(2019)}]{wang2019describing}
Wang, Q.; and Chan, A.~B. 2019.
\newblock Describing like humans: on diversity in image captioning.
\newblock In \emph{Proceedings of the IEEE/CVF Conference on Computer Vision
  and Pattern Recognition}, 4195--4203.

\bibitem[{Xu et~al.(2021)Xu, Niu, Tan, Luo, Du, and Wu}]{xu2021towards}
Xu, G.; Niu, S.; Tan, M.; Luo, Y.; Du, Q.; and Wu, Q. 2021.
\newblock Towards Accurate Text-based Image Captioning with Content Diversity
  Exploration.
\newblock In \emph{Proceedings of the IEEE/CVF Conference on Computer Vision
  and Pattern Recognition}, 12637--12646.

\bibitem[{Yang, Zhang, and Cai(2020{\natexlab{a}})}]{yang2020auto}
Yang, X.; Zhang, H.; and Cai, J. 2020{\natexlab{a}}.
\newblock Auto-encoding and distilling scene graphs for image captioning.
\newblock \emph{IEEE Transactions on Pattern Analysis and Machine
  Intelligence}.

\bibitem[{Yang, Zhang, and Cai(2020{\natexlab{b}})}]{yang2020deconfounded}
Yang, X.; Zhang, H.; and Cai, J. 2020{\natexlab{b}}.
\newblock Deconfounded image captioning: A causal retrospect.
\newblock \emph{arXiv preprint arXiv:2003.03923}.

\bibitem[{Yao et~al.(2018)Yao, Pan, Li, and Mei}]{yao2018exploring}
Yao, T.; Pan, Y.; Li, Y.; and Mei, T. 2018.
\newblock Exploring visual relationship for image captioning.
\newblock In \emph{Proceedings of the European conference on computer vision
  (ECCV)}, 684--699.

\bibitem[{Yu et~al.(2019)Yu, Li, Yu, and Huang}]{yu2019multimodal}
Yu, J.; Li, J.; Yu, Z.; and Huang, Q. 2019.
\newblock Multimodal transformer with multi-view visual representation for
  image captioning.
\newblock \emph{IEEE transactions on circuits and systems for video
  technology}, 30(12): 4467--4480.

\bibitem[{Zhang et~al.(2021{\natexlab{a}})Zhang, Shi, Tang, Xiao, Yu, and
  Zhuang}]{zhang2021consensus}
Zhang, W.; Shi, H.; Tang, S.; Xiao, J.; Yu, Q.; and Zhuang, Y.
  2021{\natexlab{a}}.
\newblock Consensus graph representation learning for better grounded image
  captioning.
\newblock In \emph{Proc 35 AAAI Conf on Artificial Intelligence}.

\bibitem[{Zhang et~al.(2019)Zhang, Tang, Cao, Pu, Wu, and
  Zhuang}]{zhang2019frame}
Zhang, W.; Tang, S.; Cao, Y.; Pu, S.; Wu, F.; and Zhuang, Y. 2019.
\newblock Frame augmented alternating attention network for video question
  answering.
\newblock \emph{IEEE Transactions on Multimedia}, 22(4): 1032--1041.

\bibitem[{Zhang et~al.(2020{\natexlab{a}})Zhang, Tang, Cao, Xiao, Pu, Wu, and
  Zhuang}]{zhang2020photo}
Zhang, W.; Tang, S.; Cao, Y.; Xiao, J.; Pu, S.; Wu, F.; and Zhuang, Y.
  2020{\natexlab{a}}.
\newblock Photo Stream Question Answer.
\newblock In \emph{Proceedings of the 28th ACM International Conference on
  Multimedia}, 3966--3975.

\bibitem[{Zhang et~al.(2021{\natexlab{b}})Zhang, Tang, Su, Xiao, and
  Zhuang}]{zhang2021tell}
Zhang, W.; Tang, S.; Su, J.; Xiao, J.; and Zhuang, Y. 2021{\natexlab{b}}.
\newblock Tell and guess: cooperative learning for natural image caption
  generation with hierarchical refined attention.
\newblock \emph{Multimedia Tools and Applications}, 80(11): 16267--16282.

\bibitem[{Zhang et~al.(2020{\natexlab{b}})Zhang, Wang, Tang, Shi, Shi, Xiao,
  Zhuang, and Wang}]{zhang2020relational}
Zhang, W.; Wang, X.~E.; Tang, S.; Shi, H.; Shi, H.; Xiao, J.; Zhuang, Y.; and
  Wang, W.~Y. 2020{\natexlab{b}}.
\newblock Relational graph learning for grounded video description generation.
\newblock In \emph{Proceedings of the 28th ACM International Conference on
  Multimedia}, 3807--3828.

\bibitem[{Zhu et~al.(2017)Zhu, Park, Isola, and Efros}]{zhu2017unpaired}
Zhu, J.-Y.; Park, T.; Isola, P.; and Efros, A.~A. 2017.
\newblock Unpaired image-to-image translation using cycle-consistent
  adversarial networks.
\newblock In \emph{Proceedings of the IEEE international conference on computer
  vision}, 2223--2232.

\end{thebibliography}
\end{document}